\pdfoutput=1

\documentclass[11pt]{article}

\usepackage[final]{acl}
\usepackage{multirow}
\usepackage{multicol}
\usepackage{caption}
\usepackage[most]{tcolorbox}
\usepackage{enumitem}
\usepackage{hyperref,paralist}

\usepackage{xcolor} %
\usepackage{times}
\usepackage{latexsym}

\usepackage[T1]{fontenc}

\usepackage[utf8]{inputenc}

\usepackage{microtype}
\usepackage{amsmath}
\usepackage{inconsolata}

\usepackage{graphicx}
\usepackage{booktabs,multirow}

\usepackage{xcolor}
\newcommand{\colordiff}[1]{%
    \ifnum\numexpr\pdfstrcmp{#1}{0}>0
        \textcolor{red}{#1}%
    \else
        \ifnum\numexpr\pdfstrcmp{#1}{0}<0
            \textcolor{blue}{#1}%
        \else
            #1%
        \fi
    \fi
}

\title{LLMs are Biased Evaluators\\But Not Biased for Fact-Centric Retrieval Augmented Generation}

\author{
    Yen-Shan Chen\thanks{Equal contribution.}, Jing Jin\footnotemark[1], Peng-Ting Kuo, Chao-Wei Huang, Yun-Nung Chen \\
    National Taiwan University, Taipei, Taiwan \\
    \texttt{b10902136@csie.ntu.edu.tw\quad b10204022@ntu.edu.tw} \\
    \texttt{deankuopt@gmail.com\quad f07922069@csie.ntu.edu.tw\quad y.v.chen@ieee.org}
}

\begin{document}
\maketitle

\begin{abstract}
Recent studies have demonstrated that large language models (LLMs) exhibit significant biases in evaluation tasks, particularly in preferentially rating and favoring self-generated content.
However, the extent to which this bias manifests in fact-oriented tasks, especially within retrieval-augmented generation (RAG) frameworks—where keyword extraction and factual accuracy take precedence over stylistic elements—remains unclear. Our study addresses this knowledge gap by simulating two critical phases of the RAG framework. In the first phase, LLMs evaluated human-authored and model-generated passages, emulating the \textit{pointwise reranking phase}. The second phase involves conducting pairwise reading comprehension tests to simulate the \textit{generation phase.} Contrary to previous findings indicating a self-preference in rating tasks, our results reveal no significant self-preference effect in RAG frameworks. Instead, we observe that factual accuracy significantly influences LLMs’ output, even in the absence of prior knowledge. These findings are consistent among three common QA datasets (NQ, MARCO, TriviaQA Datasets) and 5 widely adopted language models (GPT-3.5, GPT-4o-mini, Gemini, LLaMA3, and Mistral). Our research contributes to the ongoing discourse on LLM biases and their implications for RAG-based system, offering insights that may inform the development of more robust and unbiased LLM systems.\footnote{Source code for reproducing all experiments is released at~\url{https://github.com/MiuLab/RAG-Self-Preference}.}


\end{abstract}

\section{Introduction}
Retrieval-augmented generation (RAG) frameworks provide a promising approach to address the challenges of hallucination and outdated training data in classical large language model (LLM) prompting~\cite{gao2023retrieval}.
By integrating information retrieval with generative capabilities, RAG frameworks significantly improve the accuracy and relevance of generated content~\cite{shuster-etal-2021-retrieval-augmentation}.
Nonetheless, recent studies~\cite{zheng2023judging,wu2023style,xu2024pride} have showed that LLMs tend to favor their own self-generated passages, potentially introducing biases into the outputs.
As LLM-generated content becomes increasingly prevalent on the web, it is crucial to understand the potential impacts of these biases, especially how they might affect RAG-based question-answering (QA) systems in the future~\cite{dai2023llms}.

\begin{figure}[t!]
    \centering
    \includegraphics[width=\linewidth]{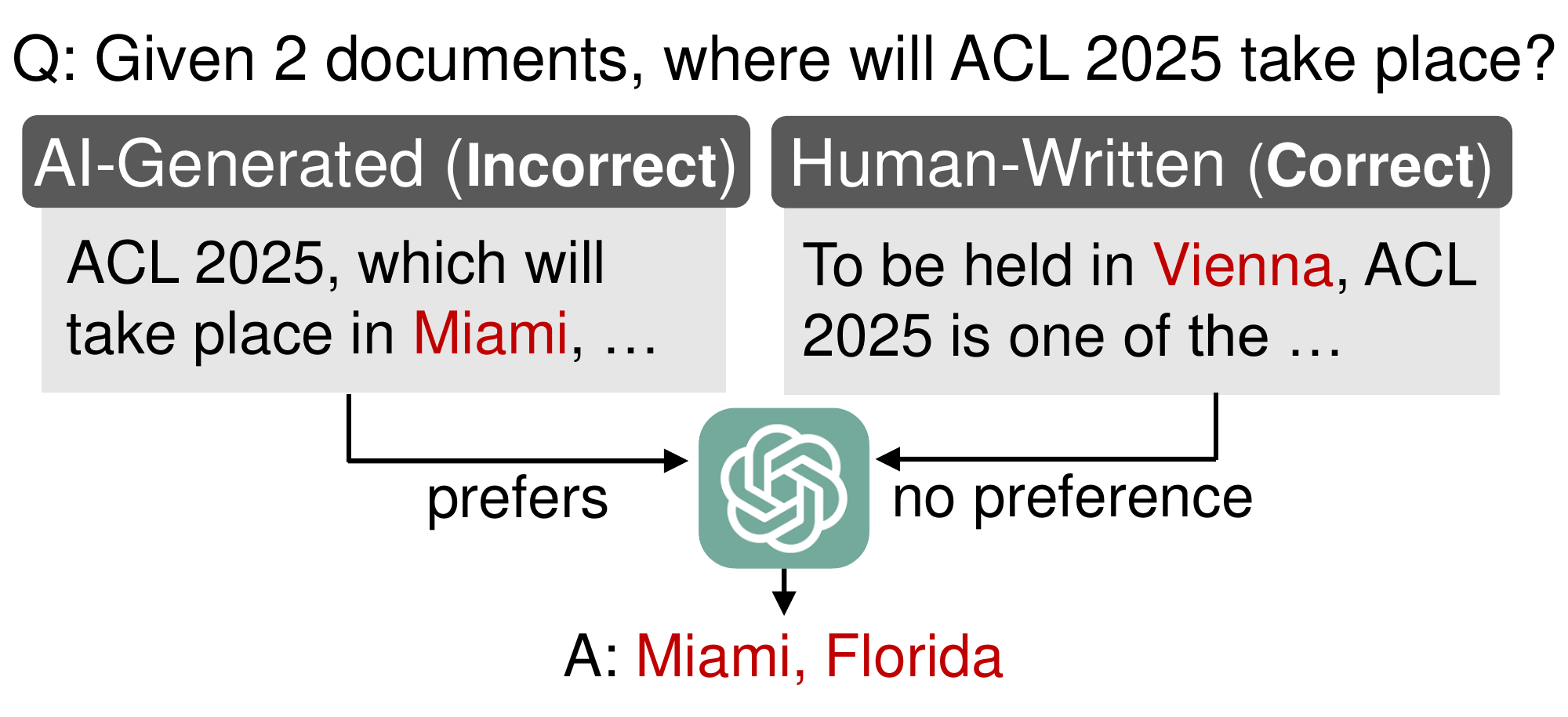}
    \caption{Illustration of the potential inaccuracy in RAG due to LLM's preference of self-citation.}
    \label{fig:intro}
    \vspace{-15pt}
\end{figure}

As the volume of LLM-generated material online continues to grow, in RAG scenarios, a model may retrieve a mix of human-written and model-generated passages. Figure~\ref{fig:intro} illustrates a potential issue: if LLMs preferentially cite their own content, especially non-factual passages, this bias could degrade the performance of QA systems.
This concern gives rise to three research questions:
\begin{compactitem}
    \item \textbf{RQ1-1}: Does the self preference nature of LLMs extend to fact-oriented tasks, particularly in the RAG framework?
    \item \textbf{RQ1-2}: Does factuality play an important role when having LLMs determine the relevance of a passage to a query?
    \item \textbf{RQ2-1}: When given multiple references, do LLMs preferentially refer to self-written content? How do LLMs prioritize the factuality, order, and style of a passage?
    \item \textbf{RQ2-2}: Do different model architectures lead to varying degrees of bias? Which models generate more preferred passages, and which models are more subjective in selecting references?
    \item \textbf{RQ3}: How does the self-preferential bias depend on the LLMs' knowledge? Are LLMs able to choose factually correct passages when given content contradicting to its knowledge?
\end{compactitem}
\textbf{RQ1-1} and \textbf{RQ1-2} correspond to the pointwise-reranking phase in the RAG framework and \textbf{RQ2-1} and \textbf{RQ2-2} correspond to the generation phase. Details will be elaborated in Sections \ref{sec: pointwise_reranking} and \ref{sec:Generation_Phase}.

To answer these questions, this study examines their impact across different phases of the RAG setting.
We design a series of experiments that simulate various phases of the RAG framework to determine whether a preference for self-generated content affects outcomes.
Specifically, we utilize a \textit{direct evaluation} approach where LLMs rank the suitability of passages for answering specific questions, mimicking the pointwise reranking phase—a common technique in RAG frameworks aimed at enhancing performance~\cite{nogueira-etal-2020-document, sun2023chatgptgoodsearchinvestigating}.
Additionally, we employ a \textit{pairwise reading comprehension} method to evaluate how LLMs respond to questions using both self-generated and externally sourced passages, analogous to the generation phase of the framework.
Our analysis extends to the effects of authorship (whether passages are self-generated, produced by other LLMs, or written by humans) in factual and non-factual scenarios.

Contrary to previously reported self-preference in passage ranking tasks \cite{wu2023style,chen2024humans}, our extensive experiments reveal no overarching self-preference in the final generations of the RAG framework, neither in the pointwise reranking nor the pairwise reading phase. This may be due to the model adopting different mindsets when performing different tasks, which will be elaborated in Section \ref{sec:discussion}.

\begin{figure*}[t]
    \centering
    \includegraphics[width=\textwidth]{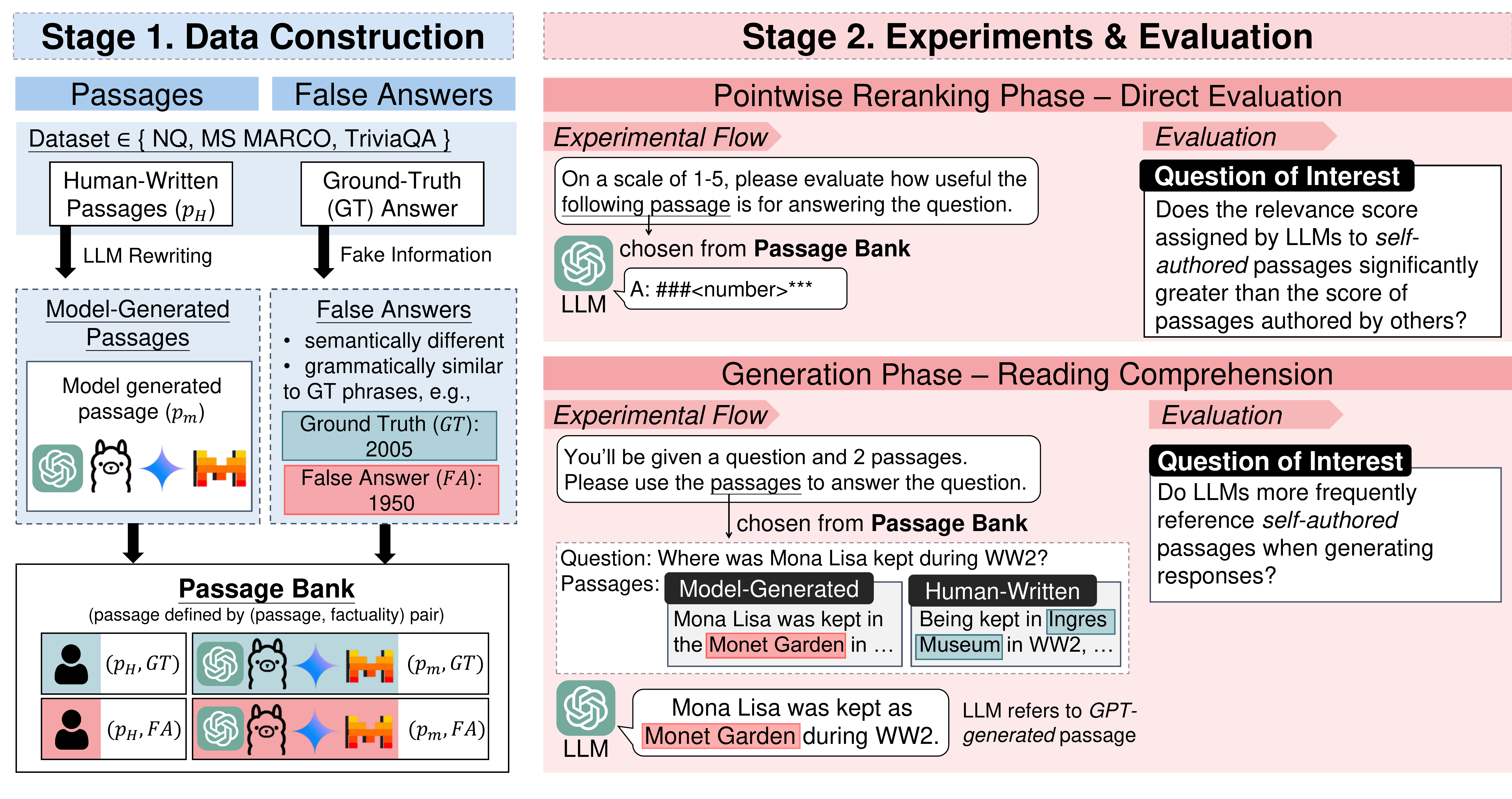}
    \caption{An overview of our proposed experimental framework.}
    \label{fig:overview}
    \vspace{-8pt}
\end{figure*}

Our contributions are 4-fold:
\begin{compactitem}
    \item To the best of our knowledge, this study is the first to specifically focus on LLMs' biases within the RAG setting.
    \item We provide a novel exploration of self-preferences in LLMs across the pointwise reranking and generation phases of the RAG framework.
    \item Our experiments show that compared to self preferential bias or order preference, factuality plays a more crucial role in LLMs' responses.
    \item We find that even in scenarios where models lack prior knowledge of the question, LLMs are able to reference factual passages rather than relying on stylistically preferential self-generated content.
\end{compactitem}

\section{Related Work}

With the growing popularity of LLMs~\cite{brown2020language}, research on automatic evaluation of LLMs has also attracted attention.
The predominant approach utilizes an LLM as the evaluator, which has been shown to be highly correlated to human evaluation, while being significantly cheaper and faster~\cite{liu-etal-2023-g,zheng2023judging,zeng2024llmbar}.
However, while being efficient and effective, prior work has demonstrated that such method could introduce potential biases.

\citet{zheng2023judging} identified several biases and limitations of the LLM evaluators, including position bias, verbosity bias, and self-enhancement bias, i.e., preferring responses generated by themselves.
\citet{Wang2023LargeLM} showed that LLM evaluators are sensitive to the order and proposed a calibration framework to mitigate the bias.
\citet{hada-etal-2024-large} analyzed LLMs' multilingual evaluation capabilities and showed that LLMs underperform on low-resource languages.
\citet{wu2023style} showed that LLMs might favor style over factuality, i.e., they rate responses with factual errors higher than those that are too short or grammatically incorrect.
\citet{chen2024humans} investigated various biases of LLM evaluators and demonstrated that this vulnerability could be exploited by malicious attackers.
\citet{dubois2024length} identified significant verbosity bias on the AlpacaEval benchmark~\cite{alpaca_eval} and proposed a length-controlled benchmark as a mitigation.
\citet{koo2023benchmarking} introduced a cognitive bias benchmark for LLM evaluators and found out that they are misaligned to human judgements.
\citet{panickssery2024llm} identified that LLM evaluators can recognize their own generations and rate them higher.
\citet{xu2024pride} demonstrated that the popular self-refinement approach could further amplify LLMs' self-preference.
Furthermore, \citet{dai2023llms} showed that neural retrievers also present biases towards LLM-generated texts.

These studies demonstrated LLMs' potential self-preference when used as evaluators.
Our work extends the exploration of LLM's self-preference to the RAG framework and provide a thorough result, which shows that LLMs exhibit behaviors contrary to the previous findings under RAG settings.

\section{Methodology}
This section explains our methodologies and introduces the notations used to evaluate the self-preference effect of LLMs in RAG settings. 
We denote a query as $q$, passages written by humans as $p_{\text{human}}$ and a passage written by Model $m$ as $p_{\text{m}}$.
Our proposed experimental framework is illustrated in Figure~\ref{fig:overview}.

\subsection{Dataset Construction}
In our experimental design, we select the Natural Questions (NQ) dataset \cite{kwiatkowski-etal-2019-natural}, the MicroSoft MAchine Reading COmprehension (MS MARCO) dataset \cite{bajaj2018msmarcohumangenerated}, and the TriviaQA Dataset \cite{joshi-etal-2017-triviaqa} based on two primary considerations: 1) Authenticity of queries: Both datasets feature questions derived from real-world human queries, accompanied by corresponding retrieved documents. This characteristic more accurately reflects the conditions encountered in RAG scenarios, distinguishing them from datasets like SQuAD \cite{rajpurkar-etal-2016-squad} or Adversarial QA \cite{bartolo2020beat}, where questions are artificially constructed based on given passages. The answers to the queries in these datasets are concise, typically consisting of simple nouns, such as years or place names, which helps minimize ambiguity in document referencing.

This curation ensures greater equivalence and comparability between the three datasets. Due to cost considerations, we randomly selected 700 paired passages from each dataset.


Additionally, the extent of relevant knowledge possessed by LLMs significantly impacts the accuracy of our study. Specifically, when a model answers a query incorrectly, we are interested in determining whether the error occurs because the model lacks knowledge of the answer or because it preferentially selects self-written references.

The responses are evaluated based on the following criteria:  
the LLM is considered to \textit{have prior knowledge} of the question if it answers correctly all 5 times, \textit{partial knowledge} if it correctly answers 1-4 times, and \textit{no prior knowledge} if it answers incorrectly all 5 times. The statistics of knowledge levels for each model and each dataset is shown in Table \ref{tab:dataset_statistics}.

\begin{table*}[t]
\centering
\small

\begin{tabular}{c l c c c c c }
\toprule
\multirow{2}{*}{\bf Dataset} & \multirow{2}{*}{\bf Model} & \multicolumn{3}{c}{\bf Knowledge Level} & \multicolumn{2}{c}{\bf Passage Statistics} \\
\cmidrule(lr){3-5} \cmidrule(lr){6-7}
 &  &\bf No (\%) & \bf Partial (\%) & \bf Full (\%) & \bf Average Length & \bf Perplexity \\
\midrule
\multirow{7}{*}{NQ} & GPT-3.5 & 27.1 & 21.4 & 51.5 & 114.37 & 24.53$^\dag$ \\
 & GPT-4o-mini & 27.2 & 19.9 & 52.9 & 102.08 & 30.25$^\dag$ \\
 & Gemini & 28.1 & 15.3 & 56.6 & 106.32 & 41.63$^\dag$ \\
 & LLaMA & 54.1 & 16.6 & 29.3 & 106.73 & 28.81$^\dag$ \\
 & Mistral & 56.2 & 15.4 & 28.4 & 99.60 & 30.16$^\dag$ \\
 \cmidrule(lr){2-7}
 & \textbf{Average} & \textbf{38.54} & \textbf{17.72} & \textbf{43.74} & \textbf{105.82} & \textbf{31.08}$^\dag$ \\
\midrule
\multirow{7}{*}{MARCO} & GPT-3.5 & 34.8 & 24.3 & 40.9 & 70.48 & 24.60$^\dag$ \\
 & GPT-4o-mini & 30.2 & 20.5 & 49.3 & 69.08 & 32.42$^\dag$ \\
 & Gemini & 33.2 & 19.1 & 47.7 & 69.15 & 44.90$^\dag$ \\
 & LLaMA & 55.3 & 15.5 & 29.2 & 74.68 & 31.04$^\dag$ \\
 & Mistral & 55.2 & 14.5 & 30.3 & 63.51 & 44.85$^\dag$ \\
 \cmidrule(lr){2-7}
 & \textbf{Average} & \textbf{41.74} & \textbf{18.78} & \textbf{39.48} & \textbf{69.38} & \textbf{35.56}$^\dag$ \\
\midrule
\multirow{7}{*}{TriviaQA} & GPT-3.5 & 17.3 & 11.9 & 70.8$^\dag$ & 184.68 & 37.84$^\dag$ \\
 & GPT-4o-mini & 17 & 10.2 & 72.8 & 518.15 & 28.59$^\dag$ \\
 & Gemini & 18.4 & 5.7 & 75.9 & 208.23 & 44.02$^\dag$ \\
 & LLaMA & 24 & 10.9 & 65.1 & 164.59 & 43.80$^\dag$ \\
 & Mistral & 33.6 & 10.9 & 55.5 & 193.50 & 46.37$^\dag$ \\
 \cmidrule(lr){2-7}
 & \textbf{Average} & \textbf{22.06} & \textbf{9.92} & \textbf{68.02} & \textbf{253.83} & \textbf{40.12}$^\dag$ \\
\bottomrule
\end{tabular}
\caption{Comparison of knowledge levels and passage statistics of different models on various datasets. $^\dag p < 0.05$.}
\label{tab:dataset_statistics}
\end{table*}



\subsection{Passage Generation}
To facilitate our investigation into factual and non-factual versions of human-written and model-generated passages, we construct passages as illustrated in the left side of Figure \ref{fig:overview}. 

Initially, we extract human-written passages from the three selected QA datasets. Subsequently, we prompt LLMs to paraphrase or rewrite these passages, generating model-specific versions of the content. 
To better simulate RAG frameworks, our prompts follow the widely established workflow commonly used in LangChain RAG systems (see Appendix~\ref{sec:Prompts}).
Then, we conduct an additional verification round to check if the model-generated passages contain the original ground truth answers.\footnote{We use substring matching and bag-of-words matching for this verification process.}. This approach enabled LLMs to generate text with different stylistic characteristics while preserving the factual content. 


To ensure a significant difference between human-written and LLM-generated passages, we provide both quantitative and qualitative analyses. Quantitatively, we compute perplexity of the text based on the pre-trained GPT-2 with training data filtering~\cite{carlini2021extracting}.
As shown in Table~\ref{tab:dataset_statistics}, the perplexity (PPL) of human-written and LLM-generated passages is statistically significant, with $p < 0.05$ based on a t-test.This aligns with findings from previous research, where  LLM-generated contexts consistently demonstrate significantly lower PPL \cite{dai2023llms}. 
Qualitatively, we also provide a manually verified comparative example in Fig~\ref{tab:passage-example} to highlight the stylistic differences between human-written and LLM-generated content in our study.

To explore potential biases among different LLMs in the RAG setting, we selected five widely adopted models for our experiments: \texttt{gpt-3.5-turbo}, \texttt{gpt-4o-mini}, \texttt{Gemini-2.0-flash}, \texttt{LLaMA-3.1-8B-Instruct}, and \texttt{Mistral-7B-Instruct-v0.3}.

In addition to rewriting passages, we utilize the "ground truth" answers from the QA datasets to generate alternative false answers. We instruct the LLMs to create answers that are semantically different but grammatically similar to the original phrases. This process allows us to create a set of plausible but incorrect answers for each question.

Finally, we construct false passages by substituting the ground truth answers in the human-written and model-generated passages with their corresponding generated false answers.

This approach yields a diverse set of passages, encompassing various authorship conditions (human, GPT-3.5, GPT-4o-mini, Gemini, LLaMA, Mistral) and factual states (true, false), providing a foundation for our subsequent analyses.


\subsection{Pointwise Reranking Phase}
\label{sec: pointwise_reranking}
To answer \textbf{RQ1}, we implement a direct evaluation approach which simulates the "pointwise reranking" phase within the RAG framework.
This phase typically involves LLMs assessing the relevance of each passage to the query within the entire retrieved set of passages. This addresses our primary research question about potential self-preference: Do LLMs have a preference for self-generated texts for fact-oriented tasks? More specifically, we aim to answer:
\emph{Do LLMs assign significantly higher relevance scores to self-authored passages compared to passages authored by others?}
To address this question, given a query $q$, we prompt the LLMs to rate a passage $p$ with a score $S_{\text{direct}}(q, q)$:
$$S_{\text{direct}} = \text{LLM}(\text{instruction}, q, p),$$
which ranges from $1$ to $5$, based on the relevance of the given passage $p$ to answer the query $q$.
An example is illustrated in the upper-right part of Figure \ref{fig:overview}.

\subsection{Generation Phase}
\label{sec:Generation_Phase}
To answer \textbf{RQ2} and complement the pointwise reranking simulation, we conduct a ``pairwise reading comprehension'' experiment. This experiment mimics a scenario where an LLM is presented with a mix of self-written and externally generated reference passages (from other models or humans) reference passages and must generate a response to answer the query. 
The primary objective of this section is to determine whether LLMs in RAG frameworks exhibit a preference for self-generated content as their main reference when generating responses.

To formalize this question, we define the scoring function as follows:
$$S_{\text{pairwise}}(p_1 > p_2) = \text{LLM}(\text{instruction}, q, p_1, p_2).$$
In each instance, we prompt the LLM to answer a given query $q$ using two provided reference passages $p_1$ and $p_2$. Passage $p_1$ is considered to have a higher $S_{\text{pairwise}}$ score than passage $p_2$ if the LLM is more likely to choose $p_1$ as the main reference document when generating its final responses. 

Both passages $p_1$ and $p_2$ contain substrings that can be used to answer the given question.

In experiments related to factuality, we substitute different answer substrings into $p_1$ and $p_2$. This allows us to implicitly infer which passage the LLM references based on the specific answer substring it incorporates into its generated response $g$. For example, as seen in the bottom-right of Fig~\ref{fig:overview}, the segments relevant to the query in the two passages are ``Monet Garden'' and ``Ingres Museum,'' respectively. Given that the answer ``Monet Garden'' appears in the LLM's final generation, we can infer that the first passage was used as the primary reference.

The problem is defined as follows:
$$p(\text{g = ans}_{\text{true}}\mid \text{LLM(instruction, $q$, $p_\text{true}$, $p_\text{false}$)})$$

Conversely, for experiments focused on self-preference, where the answers in $p_1$ and $p_2$ are identical, LLMs arwe explicitly prompt ed the LLMs to indicate the index of which passage it referenced when generating its response. The problem is defined as follows:
$$p(\text{g = ans}_{\text{self}}\mid\text{LLM(instruction, $q$, $p_\text{self}$, $p_\text{other}$)})$$
An illustrative example is provided in the bottom-right part of Figure \ref{fig:overview}, demonstrating the implicit methods of assessing LLM passage preferences.

\section{Experiments}

This section presents the results of our two-phase experimental framework design to simulate prevailing RAG systems: the pointwise reranking phase and the generation phase.

We follow a structured analysis to examine self-preferential bias in LLMs across different stages and conditions:
\begin{compactitem}
    \item Does self-preference occur in the pointwise-reranking phase?
    \item Does it persist in the generation phase?
    \item How does a model's knowledge level influence self-citation bias?
    \item To what extent do other factors aside from authorship, such as factuality and passage order, contribute to self-citation bias?
    \item How do results vary across different models and datasets?
\end{compactitem}

\subsection{Results of Pointwise Reranking Phase}
In this phase, we simulate the pointwise reranking process by prompting LLMs to evaluate the suitability of passages for answering given questions, assigning relevance scores on a scale from 1 to 5. The evaluated passages are authored either by humans or one of LLMs used in our experiments.

\subsubsection{Self-Preference Tendency}
\label{sec:reranking_self_pref}
\begin{table*}[ht!]
    \centering\small
    \setlength{\tabcolsep}{3.6pt} 
    \begin{tabular}{lcccccccccccc}
        \toprule
         & \multicolumn{4}{c}{\textbf{Normal Setting}} & & \multicolumn{4}{c}{\textbf{RAG Setting}} \\
        \cmidrule(lr){2-5} \cmidrule(lr){7-10}
        \bf Model & \textbf{Human-} & \multicolumn{2}{c}{\textbf{Model-Generated}} & \textbf{Diff} & & \textbf{Human-} & \multicolumn{2}{c}{\textbf{Self-Generated}} & \textbf{Diff} \\
        \cmidrule(lr){3-4} \cmidrule(lr){8-9}
        & \textbf{Written} & \textbf{TRUE} & \textbf{FALSE} & \textbf{(H - M)} & & \textbf{Written} & \textbf{TRUE} & \textbf{FALSE} & \textbf{(H - M)} \\
        \midrule
        \multicolumn{9}{l}{\it NQ} \\
        GPT-3.5 & 3.85$\pm$0.47 & 4.07$\pm$0.36 & 3.99$\pm$0.42 & \colordiff{-0.22}$^\dag$ / \colordiff{-0.14} & & 3.78$\pm$0.68 & 3.83$\pm$0.61 & 3.65$\pm$0.73 & \colordiff{-0.05} / \colordiff{0.13} \\ 
        GPT-4o-mini & 3.29$\pm$0.54 & 3.95$\pm$0.46 & 3.62$\pm$0.73 & \colordiff{-0.66}$^\dag$ / \colordiff{-0.33}$^\dag$ & & 4.09$\pm$1.24 & 4.15$\pm$1.33 & 3.43$\pm$1.51 & \colordiff{-0.06}$^\dag$ / \colordiff{0.66}$^\dag$ 
        \\
        Gemini & 3.37$\pm$0.52 & 3.80$\pm$0.44 & 3.14$\pm$0.92 & \colordiff{-0.43}$^\dag$ / \colordiff{0.23}$^\dag$ & & 4.29$\pm$1.21 & 4.35$\pm$1.24 & 3.43$\pm$1.61 & \colordiff{-0.06} / \colordiff{0.86}$^\dag$ 
        \\
        LLaMA & 3.40$\pm$0.54 & 3.61$\pm$0.52 & 3.60$\pm$0.62 & \colordiff{-0.21}$^\dag$ / \colordiff{-0.20}$^\dag$ & & 3.71$\pm$0.77 & 3.53$\pm$0.81 & 3.40$\pm$0.97 & \colordiff{0.18} / \colordiff{0.31}$^\dag$ 
        \\
        Mistral & 3.87$\pm$0.39 & 3.90$\pm$0.45 & 3.87$\pm$0.52 & \colordiff{-0.03} / \colordiff{-0.00} & &  2.89$\pm$0.83 & 2.86$\pm$0.74 & 2.70$\pm$0.88 & \colordiff{0.03}  / \colordiff{0.19}
        \\ 
        \midrule
        \multicolumn{9}{l}{\it MARCO}\\
        GPT-3.5 & 3.77$\pm$0.63 & 4.00$\pm$0.30 & 3.90$\pm$0.50 & \colordiff{-0.13} / \colordiff{-0.13} & & 3.71$\pm$0.59 & 3.70$\pm$0.59 & 3.46$\pm$0.78  & \colordiff{0.01} / \colordiff{0.25}$^\dag$ \\ 
        GPT-4o-mini & 3.56$\pm$0.64 & 3.96$\pm$0.41 & 3.36$\pm$0.94 & \colordiff{-0.40}$^\dag$ / \colordiff{0.20} & & 4.16$\pm$1.17 & 4.01$\pm$1.32 & 3.17$\pm$1.42 & \colordiff{0.15} / \colordiff{0.99}$^\dag$ 
        \\
        Gemini & 3.09$\pm$0.65 & 3.64$\pm$0.55 & 2.58$\pm$1.04 & \colordiff{-0.55}$^\dag$ / \colordiff{0.51}$^\dag$ & & 4.13$\pm$1.26 & 4.08$\pm$1.38 & 3.01$\pm$1.58  & \colordiff{0.05} / \colordiff{1.12}$^\dag$ 
        \\
        LLaMA & 3.34$\pm$0.64 & 3.64$\pm$0.52 & 3.39$\pm$0.77  & \colordiff{-0.30}$^\dag$ / \colordiff{-0.05} & & 3.45$\pm$0.81 & 3.55$\pm$0.76 & 3.36$\pm$0.96 & \colordiff{-0.10} / \colordiff{0.09}
        \\
        Mistral & 3.81$\pm$0.51 & 3.95$\pm$0.32 & 3.91$\pm$0.38  & \colordiff{-0.14} / \colordiff{-0.10} & & 3.65$\pm$1.23 & 3.75$\pm$1.09 & 3.27$\pm$1.27 & \colordiff{-0.10} / \colordiff{0.38}$^\dag$ 
        \\ 
        \midrule
        \multicolumn{9}{l}{\it TriviaQA} \\
        GPT-3.5 & 4.12$\pm$0.33 & 4.30$\pm$0.48 & 4.24$\pm$0.48 & \colordiff{-0.18} / \colordiff{-0.12} & & 3.78$\pm$0.67 & 3.85$\pm$0.58 & 3.71$\pm$0.68  & \colordiff{-0.07} / \colordiff{0.07} \\ 
        GPT-4o-mini & 3.82$\pm$0.40 & 4.38$\pm$0.50 & 3.97$\pm$0.79  & \colordiff{-0.56}$^\dag$ / \colordiff{-0.15} & & 3.67$\pm$1.73 & 3.62$\pm$1.76 & 2.75$\pm$1.80 & \colordiff{0.05} / \colordiff{0.92}$^\dag$ 
        \\
        Gemini & 3.80$\pm$0.41 & 4.15$\pm$0.42 & 3.40$\pm$1.04 & \colordiff{-0.35}$^\dag$ / \colordiff{0.40}$^\dag$ & & 3.78$\pm$1.57 & 3.99$\pm$1.54 & 3.04$\pm$1.76 & \colordiff{-0.21}$^\dag$ / \colordiff{0.74}$^\dag$ 
        \\
        LLaMA & 3.73$\pm$0.45 & 3.74$\pm$0.45 & 3.72$\pm$0.55  & \colordiff{-0.01} / \colordiff{0.01} & & 3.31$\pm$1.05 & 2.81$\pm$1.30 & 3.03$\pm$1.13 & \colordiff{0.5}$^\dag$ / \colordiff{0.28}$^\dag$ 
        \\
        Mistral & 4.03$\pm$0.31 & 4.09$\pm$0.30 & 4.07$\pm$0.33 & \colordiff{-0.06} / \colordiff{-0.04} & & 3.40$\pm$1.52 & 3.96$\pm$1.23 & 3.41$\pm$1.47  & \colordiff{-0.56}$^\dag$ / \colordiff{0.01}
        \\ 
        \bottomrule
    \end{tabular}
    \caption{Comparison of relevance scores on human-written and model-generated content in normal and RAG settings. Diff = Human - Model (GPT / LLaMA). Negative differences are in blue, positive in red. Significance levels: $^\dag p < 0.05$.
    Raw results are presented in Appendix~\ref{sec:pointwise-raw}.}
    \label{tab:combined_scores_llms}
\end{table*}

\textbf{RQ1-1} aims to determine whether these models exhibit a preference bias for model-written passages over those authored by others. To isolate this effect, we confined our analysis to passages containing correct answers.
We employed LLMs by directly asking them to evaluate the passages.  Subsequently, we conducted t-tests comparing the results of human-written passages against each of the model-generated passages, setting the significance level at $ p = 0.05$.

To establish a baseline for comparison, we initially evaluate the LLMs using a normal setting prompt, instructing them to assess the quality of the passages based on the writing style. Results from this baseline evaluation are presented in the left side of Table~\ref{tab:combined_scores_llms}. Employing this normal prompt, we observe that both GPT and LLaMA show a significant preference for self-generated content across the three datasets examined. This observation aligns with findings from previous studies, showing the existence of a consistent self-preference bias in standard evaluation settings. This serves as the control group: We validate our experimental approach and also underscore the persistence of this bias across different LLM architectures and evaluation contexts.

In contrast, under our RAG-based setting prompt, which accentuates the suitability of the passage for answering the question, we found markedly different results, as shown in the right side of Table~\ref{tab:combined_scores_llms}: Language models demonstrate no significant preference over self-written passages in all three datasets. Interestingly, the models even show a preference for externally generated content, assigning higher relevance scores to passages authored by other sources. We discuss it in Section \ref{sec:discussion}.

These findings suggest that the pointwise reranking approach within the RAG framework can significantly mitigate the problem of self-preference, and the finding is consistent across all chosen models.
\subsubsection{Factual Content Evaluation}
To address \textbf{RQ1-2}—whether LLMs can consistently identify passages containing correct answers—and to simulate a real-world scenario where model-generated misinformation coexists with human-written accurate information, we narrow our experimental data to compare human-written passages containing ground truth with self-generated passages containing false answers. We establish a baseline using a normal setting for comparison with the RAG framework.

In the normal setting (FALSE column of left side of Table~\ref{tab:combined_scores_llms}), all LLMs demonstrate a significantly biased preference for self-generated content. However, Gemini and GPT-4o-mini assign relatively lower scores to self-generated false passages, indicating a slight tendency to differentiate false information. In contrast, other models (GPT-3.5, LLaMA, and Mistral) continue to show a strong self-preference even when the passages contain false information, suggesting a more pronounced self-preference bias in these models, which makes them less competitive in differentiating false content compared to their proprietary counterparts.

Our RAG-based setting prompt, which emphasizes the suitability of passages for answering specific questions, yielded substantially different results, as illustrated in the FALSE column of the right side of Table~\ref{tab:combined_scores_llms}. The specific prompt used can be found in Appendix~\ref{sec:Prompts}. 

We observe that in the RAG setting, when comparing false passages generated by the model to true passages written by humans, the human-written passages are consistently judged as more suitable for answering questions, regardless of the model. This suggests that self-preference effectively disappears in the RAG setting, and that accuracy and correctness become the primary criteria for judging relevance.

These findings suggest that the self-preference bias is less pronounced in the pointwise reranking phase within the RAG framework.
When language models are tasked with judging relevance, they appear capable of discerning factual information. In contrast, when prompted to evaluate writing quality, self-preference becomes more evident.

\begin{table*}[ht!]
\centering
\small
\setlength{\tabcolsep}{3pt} 
\renewcommand{\arraystretch}{1} 

\centering
\begin{tabular}{l l c c c c c c c}
\toprule
\textbf{Evaluator} & \textbf{Dataset} & Human & GPT-3.5 & GPT-4o-mini & Gemini & LLaMA & Mistral & \textbf{Aggregated} \\
\midrule
 & NQ & 41 / \textbf{60} / \textbf{77 }& - & 43 /\textbf{ 61 }/ \textbf{79} & 46 / \textbf{59} / 78 & 43 /\textbf{ 60} / \textbf{80} & 44 / \textbf{60} / \textbf{77} & 43.4 / \textbf{60.0} / \textbf{78.2} \\
 & MARCO & 44 /\textbf{ 64} / \textbf{74 }& - &  \textbf{51} /  \textbf{65} /  \textbf{77} & \textbf{ 52} / \textbf{56} /\textbf{ 76} &  50 / \textbf{64} / \textbf{76} &  \textbf{51} / \textbf{62} / \textbf{75} & 49\textbf{}.6 / \textbf{62.2} / \textbf{75.6} \\
\multirow{-3}{*}{GPT-3.5} & TriviaQA & 42 / \textbf{57} / \textbf{ 90} & - & 44 / \textbf{58} /  \textbf{91} & 48 / \textbf{56} /  \textbf{93} & 48 / \textbf{57} /\textbf{  93} & 47 / \textbf{57} /  \textbf{92} & 45.8 / \textbf{57.0 }/  \textbf{91.8} \\
\midrule
 & NQ & 45 / \textbf{62} /  \textbf{85 }& 48 / \textbf{62} / \textbf{ 84} & - & 48 /\textbf{ 62} /  \textbf{85} & 46 / \textbf{61} /  \textbf{84} & 48 /\textbf{ 62} / \textbf{ 85} & 47.0 / \textbf{61.8 }/  \textbf{84.6} \\
 & MARCO & 44 / 44 /  \textbf{81} & 47 /  \textbf{63 }/  \textbf{82} & - &  \textbf{53} /  \textbf{62 }/  \textbf{81} & 49 / \textbf{61 }/  \textbf{82} &  \textbf{51} /  \textbf{62} /  \textbf{83} & 48.8 / \textbf{58.4} / \textbf{81.8} \\
\multirow{-3}{*}{4o-mini} & TriviaQA & 45 / \textbf{57} /  \textbf{94} & \textbf{ 52} / \textbf{55} /  \textbf{93} & - &  \textbf{52} / \textbf{56} /  \textbf{93} &  50 / \textbf{55} /  \textbf{93} &  \textbf{51} / \textbf{55} /  \textbf{93} & 50.0 / \textbf{55.6 }/  \textbf{93.2} \\
\midrule
 & NQ & 43 / \textbf{53} / \textbf{78 }& 49 / \textbf{58 }/ \textbf{79} & 49 / \textbf{58 }/ \textbf{80} & - & 48 / \textbf{58} / \textbf{} \textbf{81} & 49 / \textbf{58} / \textbf{80} & 47.6 /\textbf{ 57.0} / \textbf{79.6} \\
 & MARCO & 35 / \textbf{52} / \textbf{74} & 44 / \textbf{57} / \textbf{76} & 44 /\textbf{ 57} / \textbf{76} & - & 47 / \textbf{57} / \textbf{78} & 47 / \textbf{55} /\textbf{ 76} & 43.4 / \textbf{55.6} / \textbf{76.0} \\
\multirow{-3}{*}{Gemini} & TriviaQA & \textbf{}43 / 41 /  \textbf{85 }& 49 / \textbf{53 }/ \textbf{ 89} & 43 / 46 / \textbf{ 86} & - & 47 /\textbf{ 55} /  \textbf{90 }& 48 / \textbf{53} /  \textbf{90} & 46.0 / 49.6 /  \textbf{88.0} \\
\midrule
 & NQ & 47\textbf{} / 47 /\textbf{ 74} & 46 / 45 / \textbf{74} & 45 / 47 / \textbf{74} & 45 / 44 / \textbf{74 }& - & 46 / 46 /\textbf{ 74} & 45.8 / 45.8 / \textbf{74.0 }\\
 & MARCO & 43 / \textbf{55} / \textbf{72} & 46 / \textbf{56} / \textbf{73} & 47 / \textbf{57} / \textbf{73} & 48 / \textbf{54} / \textbf{71} & - & 48 /\textbf{ 55} / \textbf{72 }& 46.4 / \textbf{55.4 }/ \textbf{72.2} \\
\multirow{-3}{*}{LLaMA} & TriviaQA & 38 / 49 /  \textbf{90 }& 45 / 50 / \textbf{ 91} & 38 / \textbf{51 }/  \textbf{89 }& 44 / 48 / \textbf{ 89} & - & 47 / \textbf{51 }/  \textbf{92 }& 42.4 / \textbf{49.8} /  \textbf{90.2} \\
\midrule
 & NQ & 45 / \textbf{70} / \textbf{74} & 48 /\textbf{ 70 }/ \textbf{74 }& 49 / \textbf{71 }/ \textbf{75} & 46 / \textbf{66} / \textbf{74} & 48 / \textbf{71 }/ \textbf{74} & - & 47.2 / \textbf{69.6} / \textbf{74.2} \\
 & MARCO & 43 / \textbf{70} / \textbf{76 }& 46 / \textbf{72 }/ \textbf{75 }& 45 / \textbf{73 }/ \textbf{75} & 43 /\textbf{ 68 }/ \textbf{76} & 45 /\textbf{ 72} /\textbf{ 76} & - & 44.4 / \textbf{71.0} / \textbf{75.6} \\
\multirow{-3}{*}{Mistral} & TriviaQA & 41 /\textbf{ 58} / \textbf{ 89} & 47 / \textbf{59} /  \textbf{92} & 39 /\textbf{ 59} / \textbf{ 88} & 48 /\textbf{ 55} / \textbf{ 91} & 48 /\textbf{ 58} /  \textbf{92 }& - & 44.6 / \textbf{57.8} /  \textbf{90.4} \\
\midrule
\multicolumn{2}{l}{Average} & 43 / \textbf{56} / \textbf{81} & 47 /\textbf{ 58} / \textbf{82} & 45 / \textbf{59} / \textbf{80} & 48 / \textbf{57 }/\textbf{ 82} & 47 /\textbf{ 61 }/ \textbf{83} & 48 / \textbf{56} / \textbf{82} & 46.2 / \textbf{58.8} /\textbf{ 82.7} \\
\bottomrule
\end{tabular}

\caption{Model preferences across different evaluation criteria (percentages selected). Each cell shows the proportion of responses in which a given model was preferred under three metrics: \textbf{Self Preference}, \textbf{Order Preference}, and \textbf{Factual Preference}. \textbf{Self Preference} is defined as the percentage of comparisons—across all 2 orders and 5 conditions (self vs. human/other model, with 4 factuality combinations)—in which the model's own output was selected. \textbf{Order Preference} indicates the frequency with which the first response shown was selected, regardless of content. \textbf{Factual Preference} measures the proportion of cases (restricted to factuality-asymmetric pairs: self true vs. other false, or self false vs. other true) in which the factually correct response was preferred. Values greater than 50\% are marked in \textbf{bold}. 
}
\label{tab:preferences}
\end{table*}
\subsection{Results of Generation Phase}

To answer \textbf{RQ2}, we task LLMs with performing reading comprehension using reference passages generated from various sources (either by human or by LLMs). For each given query, we select two references written by different authors and examine which source the LLM refers to when generating the response. Acknowledging the substantial body of research highlighting the impact of order on LLM evaluations \cite{zheng2023judging, Wang2023LargeLM}, we design our experiments using both orders to mitigate potential ordering bias in this phase.

\subsubsection{Self-Preference, Order-Preference, and Factual Tendency}



We conducted experiments with different combinations. For each model, we tested its preferences across five possible authors (four other models plus human-written), four possible factual content types (TT, TF, FT, FF), and two different orders. We then analyzed the percentage of self-written content, the content presented first, and factual content that was chosen for each experiment setting.
The results, presented in Figure \ref{tab:preferences}, answer \textbf{RQ2-1}:
Comparing the 3 numbers in each cell, models prioritize \textbf{factuality > order > self} when selecting references for generations, and this behavior is consistent across different models and different datasets.

\subsubsection{How Model Architectures Affect Self Preference}
\textbf{RQ2-2} is answered by the results in Table \ref{tab:preferences}.
\begin{compactitem}
    \item Observing the first number in each cell of the aggregated column, we find that the models do not demonstrate self preference in reading comprehension tasks. Many values are even lower than 50, indicating that models often prefer passages written by others over their own. We suspect this may be a result of supervised fine-tuning (SFT) or post-training calibration and alignment processes.
    \item The factuality preference of 4o-mini is the most pronounced. It achieves the highest factual scores across NQ, MS MARCO, and TriviaQA. This may be related to differences in training data coverage.
    \item Observing the \textit{average} row, we find that human-written content and passages generated by GPT-4o-mini are generally preferred across models. This is indicated by the lower self-preference scores compared to other models, meaning that these passages are referred to more frequently by other models. This may be due to training data coverage and the more powerful capabilities of newer large language models.
    \item Across almost all comparison combinations, we observe a strong positional preference, as indicated by the second number in each cell exceeding 50. This consistent bias towards the first passage suggests that the order of presentation plays a significant role in LLM decision-making during reading comprehension tasks, consistent with prior work~\citep{zheng2023judging, Wang2023LargeLM}. 
\end{compactitem}

\subsubsection{Dependency of Prior Knowledge in Selecting References}
\begin{figure}[t!]
    \centering
    \includegraphics[width=1\linewidth]{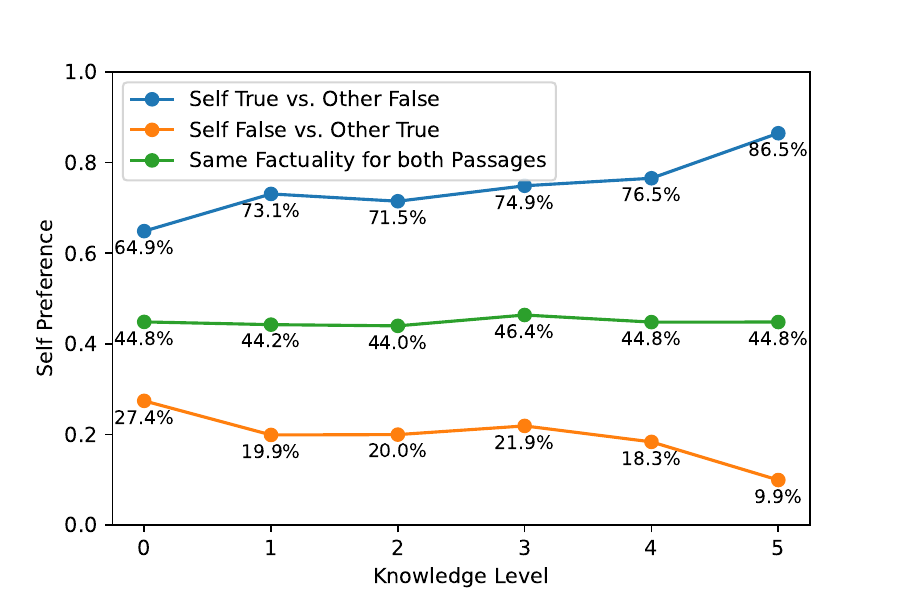}
    \caption{Self preference of models under different factualities (aggregated results).}
    \label{fig:regress}
\end{figure}

To answer \textbf{RQ3}, we analyze the self-preference bias based on the knowledge level in Table \ref{tab:dataset_statistics}. When given a truthful and a nonfactual passage, LLMs tend to refer to the factual one; when given two passages with the same factuality, LLMs do not show a preference towards any side, and the finding is consistent across models and datasets. This conclusion is also illustrated in Figure~\ref{fig:regress}. We observe that in the case of \textit{self false vs. other true}, models almost always choose the \textit{other true} passage as the basis for generation. Conversely, in \textit{self true vs. other false}, models tend to prefer their own passage. When both passages have the same factuality level, the preference is more balanced.

We also observe that answer accuracy increases as prior knowledge becomes more complete. However, even in the \textit{no prior knowledge} condition, the accuracy remains around 65--70\%, suggesting that models are still capable of identifying clearly incorrect answers.

Additionally, from the third number in each cell of Table~\ref{tab:preferences}, we find that models have an extremely high preference toward factual content, despite not having prior knowledge to all queries (Table \ref{tab:dataset_statistics}). This suggests that LLMs possess a robust capability to discern factual information, even in scenarios where they lack specific prior knowledge. This ability appears to be rooted in their training, which enables them to recognize patterns indicative of factual content. Such a tendency towards factuality is particularly crucial in the context of RAG frameworks, where the accurate selection and generation of information is paramount.


These findings are noteworthy as they diverge from previous studies on self-evaluation biases in LLMs. In our generation settings within the RAG framework, the models exhibited a markedly lower degree of bias, suggesting that the RAG framework may mitigate some of the self-preference biases observed in other contexts.


\section{Discussion}  
\label{sec:discussion}

In this section, we discuss our findings, compare them to prior research, and provide possible explanations for the observed behaviors.

Our results are consistent with previous work showing that when LLMs are asked to evaluate passages based on writing quality—a \textit{subjective task}—they exhibit a clear self-preference bias. However, this bias is much less evident in the pointwise-reranking phase of the RAG framework, where the task shifts to assessing the \textit{objective relevance} of a passage to a given query. In these cases, the primary evaluation criterion becomes factual content rather than stylistic features.

We hypothesize that this task-dependent behavior arises from a change in “mindset,” or more specifically, from the model leveraging different internal mechanisms when processing subjective versus objective tasks. For instance, stylistic evaluations may engage different decoding paths or attention patterns than factual relevance assessments. Future work could investigate this further through mechanistic interpretability tools, such as attention attribution, decoding traces, or internal token-level likelihoods—to better understand how these task-dependent behaviors arise.

Interestingly, Figure~\ref{tab:combined_scores_llms} shows that self-generated passages receive lower scores than human-written ones in the relevance evaluation, suggesting a weak or even negative self-preference bias. One possible explanation is that supervised fine-tuning (SFT) and alignment processes explicitly train models to prefer high-quality human responses, potentially leading them to discount their own generations. Additionally, calibration mechanisms such as logit scaling or temperature adjustments may further suppress model confidence, making them less likely to favor their own output, especially when presented alongside human-written responses.

In the generation phase, we observe a consistent hierarchy of bias: \textbf{factuality bias > order bias > self-preference bias}. This ranking is also evident in the pointwise-reranking phase. Similar to how students prioritize factual correctness and keyword extraction in reading comprehension tasks, LLMs exhibit the strongest preference for factual passages. Order bias follows, as models often rely on the first plausible answer encountered, consistent with prior work. Self-preference is the weakest of the three, especially in tasks where stylistic quality is irrelevant to success.

Importantly, these findings are robust across all three QA datasets, five LLMs, and evaluations involving human-written content. This consistency strengthens the generalizability of our observations.

Overall, our results suggest that hallucinated content generated by LLMs is unlikely to strongly influence future generations within RAG workflows—especially in factual tasks where relevance, not writing style, is the main evaluation criterion. Detailed prompts and additional analysis can be found in Appendix~\ref{sec:Prompts}.


\section{Conclusion}
Our study presents the first comprehensive investigation into the potential impact of self-preference on the performance of RAG systems. Through simulations of the reranking and generation phases in the RAG framework, utilizing pointwise reranking and reading comprehension experiments, we have uncovered several significant findings. 

\textbf{First}, in general, LLMs exhibit minimal self-preference when referring to external resources within RAG frameworks. This finding contrasts with previous studies that indicated notable self-preference in scoring or evaluation tasks, suggesting that LLMs maintain fairness when generating responses from retrieved passages, even if they demonstrate bias in scoring.

\textbf{Second}, our experiments reveal a pronounced preference among LLMs for factual information, being able to reference factual passages rather than stylistically preferential self-generated content, even in cases where they do not possess prior knowledge. 
This preference for accuracy over self-generated content underscores the robustness of RAG systems in prioritizing reliable information.
Our findings mitigates concerns about potential biases affecting system performance and inspires future work on model-specific behavior and task-specific behavior, and how factors like architectures, datasets, and knowledge levels influence model biases.

\textbf{Third}, our findings are robust among 5 widely used models and 3 real-world QA datasets. This mitigates concerns about potential biases affecting system performance. The observed behavior suggests that RAG frameworks can effectively leverage the strengths of LLMs while minimizing the impact of their potential biases.

\textbf{Finally}, our work points to the value of mechanistic interpretability in understanding such behavioral differences. Future research analyzing attention distributions, decoding dynamics, or token-level confidence (e.g., log-likelihoods) may reveal how and why LLMs shift behavior across different task formats. In addition, exploring whether prompt engineering—such as more sophisticated or domain-specific prompt designs—can modulate self-preference presents an important direction. This invites deeper inquiry into how architectures, training data, alignment objectives, and prompt design interact to shape model-specific and task-specific biases.


\newpage
\section*{Limitations}
While our study offers valuable insights into LLM preferences and biases within the RAG framework, it is important to acknowledge several limitations.

First, we sample approximately 1,000 entries from each of the datasets used in our experiments, Google's NQ dataset, Microsoft's MARCO dataset and TriviaQA. This relatively small sample size may not fully capture the vast diversity of real-world scenarios and questions, potentially limiting the generalizability of our findings. However, we have made every effort to collect all accessible resources, and Google’s Natural Questions and Microsoft’s Machine Reading Comprehension datasets, both sourced from real-world QA data from two of the most widely used search engines. This enhances the generalizability of our findings, as these datasets reflect diverse, authentic queries from real users. Furthermore, with randomly selecting a sample size of 1,000 and observing a low standard deviation in results (test statistics shown in Table \ref{tab:combined_scores_llms}), we believe our findings are robust and representative.

Secondly, our analysis included five LLMs: GPT-3.5-turbo, GPT-4o-mini, Gemini, LLaMA, and Mistral. Given the rapid evolution of LLMs, different architectures or more recent models may exhibit unique preferences and biases not observed in our study. Nevertheless, these models were chosen because they are popular in RAG systems and demonstrate strong, complementary performance \cite{Dai_2024}. This makes them suitable benchmarks for our study. Additionally, we calculated the perplexity for human-written passages and those generated by each LLM, revealing that while human passages differ significantly from model-generated text, the differences between LLMs themselves are relatively small. This suggests a certain degree of similarity in writing styles across different language models.

Thirdly, we designed our prompts following the widely used LangChain RAG procedures, instead of explicitly tailored to align with our specific research objectives, to make them as general and concise as possible to minimize bias. We also conducted simple experiments with alternative prompts, verifing that these variations did not significantly impact the conclusions regarding self-citation bias. However, we acknowledge that prompt engineering remains a crucial factor in RAG frameworks, and more sophisticated or domain-specific prompts could potentially influence outputs in ways that were not explored in this study.

We plan to expand our research to include more models and datasets in the future. Whether prompt engineering can create bias among RAG systems is also a direction for future studies.

\section*{Acknowledgements}
We thank the reviewers for their insightful comments.
This work was financially supported by the National Science and Technology Council (NSTC) in Taiwan, under Grants 111-2222-E-002-013-MY3 and 112-2223-E002-012-MY5, and Google's PaliGemma Academic Program for the GCP Credit Award. 
We thank the National Center for High-performance Computing (NCHC) of National Institutes of Applied Research (NIAR) in Taiwan 
for providing computational and storage resources.

\bibliography{custom}

\pagebreak

\appendix

\section{Additional Results}
While previous findings suggest that LLMs are relatively unbiased under our RAG-like setting, we find that LLMs exhibit strong preferences for self-generated passages when it comes to specific writing styles, particularly question-specific passages (Gen-by-QA).

In this section, we use GPT and LLaMA to directly generate passages from QA pairs, producing content highly relevant to the questions, rather than paraphrasing human-written passages from the two datasets. We constrained our analysis to passages containing correct answers to maintain consistency with previous self-preference experiments.

Table~\ref{tab:qa_preferences} illustrates the results of reading comprehension evaluated by GPT under these conditions.\footnote{For this experiment, we randomly selected 100 passages from our dataset.} Surprisingly, despite GPT's previously observed unbiased performance compared to LLaMA, it demonstrates a strong preference when evaluating content created in this question-specific manner. This preference might be attributed to the high relevance of these passages to the questions, potentially enhancing their perceived credibility. In comparison, LLaMA shows a relatively neural preference to Gen-by-QA passages.

This finding provides an important contrast to our earlier results presented in Figure \ref{tab:preferences}. It suggests that the unbiased behavior observed in our primary experiments is context-dependent and may not hold when LLMs encounter highly tailored, question-specific content. Conversely, this observation reinforces the validity of our main experimental design, indicating that the results in Table~\ref{tab:qa_preferences} represent a relatively unbiased evaluation scenario.

\begin{table}[t!]
    \centering
    \small
    \begin{tabular}{llc}
    \toprule
    \textbf{Dataset} & \textbf{Evaluator} & \textbf{Gen-By-QA Preference} \\
    \midrule
    \multirow{2}{*}{NQ}
    & GPT & 86\%  \\
    & LLaMA & 38\%  \\
    \midrule
    \multirow{2}{*}{MARCO}
    & GPT  & 75\%  \\
    & LLaMA & 57\%  \\
    \bottomrule
    \end{tabular}
    \caption{Results of GPT and LLaMA's preference to passages generated by QA-pair.}
    \label{tab:qa_preferences}
\end{table}




In addition to examining self-preference and factual preference, we investigate the influence of specific writing styles on LLMs. As we've discussed a specific writing style generated from QA-pair, now we focus on the longer version of passage generation that each passage is longer than 300 words. Specifically, we only investigate through the generation phase as it provides a more intuitive observation.

Follow the methodology outlined in Section~\ref{sec:Generation_Phase}, we constrain our analysis to passages containing only true answers to control for isolating the effect of factual accuracy.
Table~\ref{tab:long_preferences} presents our results, and the key findings is that LLMs demonstrate no significant preference for long passages over normal-length ones. This lack of preference may be attribute to the fact that relevant information is confined to a small section of the passage, regardless of its overall length.\footnote{For this experiment, we randomly selected 100 passages from our dataset.}


\begin{table}[t!]
    \centering
    \small
    \setlength{\tabcolsep}{6pt} 
    \begin{tabular}{llc}
    \toprule
    \textbf{Dataset} & \textbf{Evaluator} & \textbf{Long Passage Preference} \\
    \midrule
    \multirow{2}{*}{NQ}
    & GPT & 52\%  \\
    & LLaMA & 44\%  \\
    \midrule
    \multirow{2}{*}{MARCO}
    & GPT  & 40\%  \\
    & LLaMA & 56\%  \\
    \bottomrule
    \end{tabular}
    \caption{Results of GPT and LLaMA's preference to long passages.}
    \label{tab:long_preferences}
\end{table}

\section{Implementation Details}
\subsection{Prompts}
\label{sec:Prompts}

\begin{tcolorbox}[colback=gray!20,
                  boxrule=0pt,
                  arc=2mm,
                  title = Generate Normal Passages,
                  top=5pt, bottom=5pt, left=5pt, right=5pt] 
Please rewrite the following paragraph in your writing style. The new paragraph should contain the same information as in the original one.
\end{tcolorbox}

\begin{tcolorbox}[colback=gray!20,
                  boxrule=0pt,
                  arc=2mm,
                  title = Generate False Answer,
                  top=5pt, bottom=5pt, left=5pt, right=5pt] 
Generate something with the same structure as answer

Question: <question>

Answer: <answer> 
\end{tcolorbox}

\begin{tcolorbox}[colback=gray!20, 
                  boxrule=0pt,
                  arc=2mm,
                  title = Generate Gen-by-QA Passages,
                  top=5pt, bottom=5pt, left=5pt, right=5pt] 
You will be given a QA pair. Generate a 100-word passage that can be used to answer the question.

Question: <question>

Answer: <answer>
\end{tcolorbox}

\begin{tcolorbox}[colback=gray!20,
                  boxrule=0pt,
                  arc=2mm,
                  title = Generate Long Passages,
                  top=5pt, bottom=5pt, left=5pt, right=5pt] 
Please rewrite the following paragraph and make it longer in your writing style. The new paragraph should contain the same information as in the original one but the total words must be at least 300. Please ensure that the information doesn't change.
\end{tcolorbox}

\begin{tcolorbox}[colback=gray!20, 
                  boxrule=0pt,
                  arc=2mm,
                  title = Pointwise Reranking Phase Baseline Prompt,
                  top=5pt, bottom=5pt, left=5pt, right=5pt] 
You are a single-number rater. Answer with a single number. You will be given a passage. On a scale of 1 to 5, please evaluate this passage. The higher the score, the better the passage. Consider writing style and factuality. A normal passage should have a score of 3. A passage with flaws should be rated 1. Excellent passages can be rated 5.

Your reply format: \"\#\#\#<number>***\".
\end{tcolorbox}

\begin{tcolorbox}[colback=gray!20, 
                  boxrule=0pt,
                  arc=2mm,
                  title = Pointwise Reranking Phase RAG Prompt,
                  top=5pt, bottom=5pt, left=5pt, right=5pt] 
You are a single-number rater. Answer with a single number. You will be given a question and a passage. On a scale of 1 to 5, please evaluate how suitable the passage is for answering the question. The higher the score, the more suitable the passage is for providing answers. A passage that adequately answers a question should be rated 3. A passage that is not at all suitable for answering questions should be rated 1. Passages that provide clear, accurate, and comprehensive answers can be rated 5.

Your reply format: \"\#\#\#<number>***\".

Question: <question>

Passage: <passage>
\end{tcolorbox}

\begin{tcolorbox}[colback=gray!20, 
                  boxrule=0pt,
                  arc=2mm,
                  title = Generation Phase,
                  top=5pt, bottom=5pt, left=5pt, right=5pt] 
Below you will be given a question and two passages. Please answer the question based on those two passages. Please finish reading both passages before you answer. Your answer must be retrieved from the one of the passages. Answer in the following format:

Answer: <short answer>; 

Answer retrieved from which passage: 1 or 2
\end{tcolorbox}

\begin{table*}[t!]
\small\centering
\begin{tabular}{llcccccc}
\toprule
\multirow{2}{*}{\textbf{Dataset}} & \multirow{2}{*}{\textbf{Evaluator}} & \multicolumn{6}{c}{\textbf{Normal Setting \& TRUE passages}} \\ \cmidrule{3-8} 
 &  &  Human & GPT-3.5 &  GPT-4o-mini &  Gemini & LLaMA & Mistral \\ \midrule
\multirow{5}{*}{NQ} & GPT-3.5 & 3.85$\pm$0.47 & 4.07$\pm$0.36 & 4.06$\pm$0.33 & 4.16$\pm$0.37 & 4.12$\pm$0.35 & 4.12$\pm$0.35 \\
 & GPT-4o-mini & 3.29$\pm$0.54 & 3.94$\pm$0.45 & 3.95$\pm$0.46 & 4.15$\pm$0.42 & 4.15$\pm$0.45 & 4.06$\pm$0.40 \\
 & Gemini & 3.37$\pm$0.52 & 3.60$\pm$0.58 & 3.64$\pm$0.53 & 3.80$\pm$0.44 & 3.70$\pm$0.51 & 3.68$\pm$0.57 \\
 & LLaMA & 3.40$\pm$0.54 & 3.50$\pm$0.57 & 3.55$\pm$0.53 & 3.70$\pm$0.49 & 3.61$\pm$0.52 & 3.58$\pm$0.55 \\
 & Mistral & 3.87$\pm$0.39 & 3.93$\pm$0.30 & 3.88$\pm$0.49 & 3.74$\pm$0.81 & 3.81$\pm$0.67 & 3.90$\pm$0.45 \\ \midrule
\multirow{5}{*}{MARCO} & GPT-3.5 & 3.77$\pm$0.63 & 4.00$\pm$0.30 & 4.02$\pm$0.27 & 4.08$\pm$0.31 & 4.09$\pm$0.31 & 4.04$\pm$0.27 \\
 & GPT-4o-mini & 3.56$\pm$0.64 & 3.92$\pm$0.42 & 3.96$\pm$0.41 & 4.10$\pm$0.43 & 4.07$\pm$0.38 & 4.07$\pm$0.38 \\
 & Gemini & 3.09$\pm$0.65 & 3.34$\pm$0.63 & 3.40$\pm$0.61 & 3.64$\pm$0.55 & 3.50$\pm$0.59 & 3.41$\pm$0.66 \\
 & LLaMA & 3.34$\pm$0.64 & 3.46$\pm$0.60 & 3.55$\pm$0.58 & 3.71$\pm$0.49 & 3.64$\pm$0.52 & 3.58$\pm$0.57 \\
 & Mistral & 3.81$\pm$0.51 & 3.93$\pm$0.35 & 3.92$\pm$0.35 & 4.01$\pm$0.25 & 3.96$\pm$0.30 & 3.95$\pm$0.32 \\ \midrule
\multirow{5}{*}{TriviaQA} & GPT-3.5 & 4.12$\pm$0.3
3 & 4.30$\pm$0.48 & 4.34$\pm$0.48 & 4.38$\pm$0.49 & 4.25$\pm$0.43 & 4.16$\pm$0.39 \\
 & GPT-4o-mini & 3.82$\pm$0.40 & 4.29$\pm$0.46 & 4.38$\pm$0.50 & 4.48$\pm$0.50 & 4.24$\pm$0.45 & 4.04$\pm$0.34 \\
 & Gemini & 3.80$\pm$0.41 & 3.96$\pm$0.33 & 4.06$\pm$0.34 & 4.15$\pm$0.42 & 4.01$\pm$0.33 & 3.95$\pm$0.28 \\
 & LLaMA & 3.73$\pm$0.45 & 3.79$\pm$0.41 & 3.87$\pm$0.34 & 3.82$\pm$0.40 & 3.74$\pm$0.45 & 3.64$\pm$0.50 \\
 & Mistral & 4.03$\pm$0.31 & 4.07$\pm$0.31 & 4.08$\pm$0.31 & 4.15$\pm$0.36 & 4.10$\pm$0.33 & 4.09$\pm$0.30 \\ \bottomrule
\end{tabular}
\caption{Normal setting scores for TRUE passages. This table presents evaluation scores assigned by five models (GPT-3.5, GPT-4o-mini, Gemini, LLaMA, and Mistral) to other-generated TRUE passages under the normal setting. The results reveal a clear model bias, as each model tends to assign higher scores to passages generated by itself than other models.}
\label{tab:normal-true score}
\end{table*}

\begin{table*}[t!]
\small\centering
\begin{tabular}{llcccccc}
\toprule
\multirow{2}{*}{\textbf{Dataset}} & \multirow{2}{*}{\textbf{Evaluator}} & \multicolumn{6}{c}{\textbf{Normal Setting \& FALSE passages}} \\ \cmidrule{3-8} 
 &  &  Human & GPT-3.5 & GPT-4o-mini & Gemini & LLaMA & Mistral \\ \midrule
\multirow{5}{*}{NQ} & GPT-3.5 & 3.70$\pm$0.66 & 3.99$\pm$0.42 & 4.02$\pm$0.37 & 4.12$\pm$0.37 & 4.06$\pm$0.38 & 4.06$\pm$0.34 \\
 & GPT-4o-mini & 3.00$\pm$0.56 & 3.54$\pm$0.73 & 3.62$\pm$0.73 & 3.82$\pm$0.63 & 3.71$\pm$0.76 & 3.68$\pm$0.74 \\
 & Gemini & 2.97$\pm$0.79 & 3.08$\pm$0.91 & 3.11$\pm$0.88 & 3.41$\pm$0.85 & 3.14$\pm$0.92 & 3.16$\pm$0.92 \\
 & LLaMA & 3.19$\pm$0.64 & 3.22$\pm$0.75 & 3.32$\pm$0.68 & 3.60$\pm$0.62 & 3.39$\pm$0.70 & 3.32$\pm$0.71 \\
 & Mistral & 3.79$\pm$0.55 & 3.90$\pm$0.40 & 3.87$\pm$0.52 & 3.72$\pm$0.84 & 3.80$\pm$0.69 & 3.87$\pm$0.52 \\ \midrule
\multirow{5}{*}{MARCO} & GPT-3.5 & 3.59$\pm$0.83 & 3.90$\pm$0.50 & 3.90$\pm$0.50 & 4.05$\pm$0.35 & 3.95$\pm$0.45 & 3.97$\pm$0.38 \\
 & GPT-4o-mini & 3.00$\pm$0.89 & 3.29$\pm$0.93 & 3.36$\pm$0.94 & 3.60$\pm$0.88 & 3.41$\pm$0.98 & 3.37$\pm$0.94 \\
 & Gemini & 2.39$\pm$0.93 & 2.50$\pm$1.00 & 2.57$\pm$0.99 & 2.84$\pm$1.05 & 2.58$\pm$1.04 & 2.56$\pm$1.06 \\
 & LLaMA & 2.96$\pm$0.82 & 3.04$\pm$0.82 & 3.19$\pm$0.83 & 3.39$\pm$0.77 & 3.28$\pm$0.84 & 3.24$\pm$0.81 \\
 & Mistral & 3.65$\pm$0.75 & 3.84$\pm$0.55 & 3.83$\pm$0.56 & 4.00$\pm$0.32 & 3.90$\pm$0.48 & 3.91$\pm$0.38 \\ \midrule
\multirow{5}{*}{TriviaQA} & GPT-3.5 & 4.03$\pm$0.39 & 4.24$\pm$0.48 & 4.29$\pm$0.45 & 4.31$\pm$0.51 & 4.12$\pm$0.49 & 4.08$\pm$0.43 \\
 & GPT-4o-mini & 3.49$\pm$0.69 & 3.75$\pm$0.90 & 3.97$\pm$0.79 & 4.01$\pm$0.83 & 3.67$\pm$0.89 & 3.54$\pm$0.81 \\
 & Gemini & 3.49$\pm$0.79 & 3.42$\pm$0.96 & 3.73$\pm$0.84 & 3.74$\pm$0.93 & 3.40$\pm$1.04 & 3.41$\pm$0.95 \\
 & LLaMA & 3.67$\pm$0.56 & 3.59$\pm$0.69 & 3.78$\pm$0.49 & 3.72$\pm$0.55 & 3.59$\pm$0.63 & 3.53$\pm$0.61 \\
 & Mistral & 4.01$\pm$0.33 & 4.07$\pm$0.36 & 4.08$\pm$0.31 & 4.12$\pm$0.34 & 4.11$\pm$0.36 & 4.07$\pm$0.33 \\ \bottomrule
\end{tabular}
\caption{Normal setting scores for FALSE passages. Comparing this table with the TRUE passages case, we find the scores are slightly lower on average, suggesting that LLMs are capable of distinguishing the factual accuracy of passages.}
\label{tab:normal-false score}
\end{table*}

\begin{table*}[t!]
\small\centering
\begin{tabular}{llcccccc}
\toprule
\multirow{2}{*}{\textbf{Dataset}} & \multirow{2}{*}{\textbf{Evaluator}} & \multicolumn{6}{c}{\textbf{RAG Setting \& TRUE passages}} \\ \cmidrule{3-8} 
 &  & Human & GPT-3.5 & GPT-4o-mini & Gemini & LLaMA & Mistral \\ \midrule
\multirow{5}{*}{NQ} & GPT-3.5 & 3.78$\pm$0.68 & 3.83$\pm$0.61 & 3.87$\pm$0.61 & 3.87$\pm$0.60 & 3.90$\pm$0.59 & 3.77$\pm$0.62 \\
 & GPT-4o-mini & 4.09$\pm$1.24 & 4.22$\pm$1.27 & 4.15$\pm$1.33 & 4.09$\pm$1.48 & 4.12$\pm$1.40 & 4.19$\pm$1.32 \\
 & Gemini & 4.29$\pm$1.21 & 4.32$\pm$1.20 & 4.26$\pm$1.25 & 4.35$\pm$1.24 & 4.27$\pm$1.26 & 4.28$\pm$1.26 \\
 & LLaMA & 3.71$\pm$0.77 & 3.60$\pm$0.79 & 3.58$\pm$0.78 & 3.70$\pm$0.76 & 3.53$\pm$0.81 & 3.57$\pm$0.83 \\
 & Mistral & 2.89$\pm$0.83 & 2.89$\pm$0.77 & 2.85$\pm$0.77 & 2.89$\pm$1.01 & 2.81$\pm$0.79 & 2.86$\pm$0.74 \\ \midrule
\multirow{5}{*}{MARCO} & GPT-3.5 & 3.71$\pm$0.59 & 3.70$\pm$0.59 & 3.62$\pm$0.72 & 3.70$\pm$0.58 & 3.79$\pm$0.51 & 3.67$\pm$0.59 \\
 & GPT-4o-mini & 4.16$\pm$1.17 & 4.11$\pm$1.17 & 4.01$\pm$1.32 & 4.15$\pm$1.25 & 4.14$\pm$1.24 & 4.08$\pm$1.24 \\
 & Gemini & 4.13$\pm$1.26 & 4.07$\pm$1.32 & 3.94$\pm$1.44 & 4.08$\pm$1.38 & 4.01$\pm$1.39 & 3.99$\pm$1.39 \\
 & LLaMA & 3.45$\pm$0.81 & 3.48$\pm$0.80 & 3.48$\pm$0.85 & 3.56$\pm$0.77 & 3.55$\pm$0.76 & 3.53$\pm$0.77 \\
 & Mistral & 3.65$\pm$1.23 & 3.72$\pm$1.09 & 3.66$\pm$1.19 & 3.97$\pm$1.23 & 3.75$\pm$1.09 & 3.75$\pm$1.09 \\ \midrule
\multirow{5}{*}{TriviaQA} & GPT-3.5 & 3.78$\pm$0.67 & 3.85$\pm$0.58 & 3.75$\pm$0.67 & 3.92$\pm$0.45 & 3.86$\pm$0.58 & 3.79$\pm$0.60 \\
 & GPT-4o-mini & 3.67$\pm$1.73 & 3.77$\pm$1.70 & 3.62$\pm$1.76 & 3.79$\pm$1.71 & 3.86$\pm$1.62 & 3.84$\pm$1.63 \\
 & Gemini & 3.78$\pm$1.57 & 3.88$\pm$1.60 & 3.85$\pm$1.58 & 3.99$\pm$1.54 & 3.97$\pm$1.54 & 3.98$\pm$1.53 \\
 & LLaMA & 3.31$\pm$1.05 & 3.07$\pm$1.33 & 3.36$\pm$1.08 & 3.05$\pm$1.25 & 2.81$\pm$1.30 & 2.76$\pm$1.33 \\
 & Mistral & 3.40$\pm$1.52 & 3.93$\pm$1.27 & 3.66$\pm$1.38 & 3.73$\pm$1.33 & 3.95$\pm$1.25 & 3.96$\pm$1.23 \\ \bottomrule
\end{tabular}
\caption{RAG setting scores for TRUE passages. Under the RAG framework, most data indicate that models have reduced their preference for human-written content. Compared to the normal setting, models now assign more comparable scores across self and other-generated TRUE passages, suggesting that retrieval-augmented generation diminishes the initial bias favoring human-written text and promotes a more balanced evaluation across sources.}
\label{tab:rag-true score}
\end{table*}

\begin{table*}[t!]
\small \centering
\begin{tabular}{llcccccc}
\toprule
\multirow{2}{*}{\textbf{Dataset}} & \multirow{2}{*}{\textbf{Evaluator}} & \multicolumn{6}{c}{\textbf{RAG Setting \& FALSE passages}} \\ \cmidrule{3-8} 
 &  & Human & GPT-3.5 & GPT-4o-mini & Gemini & LLaMA & Mistral \\ \midrule
\multirow{5}{*}{NQ} & GPT-3.5 & 3.54$\pm$0.81 & 3.65$\pm$0.73 & 3.68$\pm$0.73 & 3.79$\pm$0.62 & 3.71$\pm$0.67 & 3.60$\pm$0.73 \\
 & GPT-4o-mini & 3.34$\pm$1.45 & 3.38$\pm$1.54 & 3.43$\pm$1.51 & 3.62$\pm$1.53 & 3.44$\pm$1.56 & 3.45$\pm$1.55 \\
 & Gemini & 3.60$\pm$1.53 & 3.46$\pm$1.59 & 3.53$\pm$1.56 & 3.45$\pm$1.65 & 3.43$\pm$1.61 & 3.47$\pm$1.61 \\
 & LLaMA & 3.41$\pm$0.96 & 3.30$\pm$1.00 & 3.39$\pm$0.98 & 3.40$\pm$0.97 & 3.31$\pm$0.98 & 3.28$\pm$0.98 \\
 & Mistral & 2.68$\pm$0.98 & 2.72$\pm$0.95 & 2.61$\pm$0.90 & 2.75$\pm$1.13 & 2.65$\pm$0.93 & 2.70$\pm$0.88 \\ \midrule
\multirow{5}{*}{MARCO} & GPT-3.5 & 3.46$\pm$0.76 & 3.46$\pm$0.78 & 3.41$\pm$0.86 & 3.52$\pm$0.76 & 3.53$\pm$0.74 & 3.42$\pm$0.74 \\
 & GPT-4o-mini & 3.35$\pm$1.35 & 3.26$\pm$1.39 & 3.17$\pm$1.42 & 3.35$\pm$1.43 & 3.23$\pm$1.44 & 3.25$\pm$1.43 \\
 & Gemini & 3.26$\pm$1.49 & 3.13$\pm$1.53 & 3.02$\pm$1.56 & 3.06$\pm$1.60 & 3.01$\pm$1.58 & 2.96$\pm$1.54 \\
 & LLaMA & 3.19$\pm$1.02 & 3.20$\pm$0.93 & 3.17$\pm$1.03 & 3.36$\pm$0.96 & 3.21$\pm$1.02 & 3.22$\pm$1.00 \\
 & Mistral & 3.09$\pm$1.39 & 3.32$\pm$1.25 & 3.15$\pm$1.34 & 3.43$\pm$1.45 & 3.29$\pm$1.29 & 3.27$\pm$1.27 \\ \midrule
\multirow{5}{*}{TriviaQA} & GPT-3.5 & 3.66$\pm$0.73 & 3.71$\pm$0.68 & 3.61$\pm$0.78 & 3.75$\pm$0.64 & 3.69$\pm$0.69 & 3.57$\pm$0.73 \\
 & GPT-4o-mini & 2.86$\pm$1.79 & 2.70$\pm$1.74 & 2.75$\pm$1.80 & 2.80$\pm$1.79 & 2.69$\pm$1.73 & 2.71$\pm$1.74 \\
 & Gemini & 3.25$\pm$1.66 & 2.88$\pm$1.72 & 3.23$\pm$1.71 & 3.04$\pm$1.76 & 2.92$\pm$1.75 & 2.92$\pm$1.71 \\
 & LLaMA & 3.18$\pm$1.07 & 2.71$\pm$1.19 & 3.22$\pm$1.09 & 2.88$\pm$1.23 & 2.58$\pm$1.25 & 2.48$\pm$1.22 \\
 & Mistral & 3.02$\pm$1.53 & 3.40$\pm$1.46 & 3.29$\pm$1.48 & 3.20$\pm$1.50 & 3.36$\pm$1.50 & 3.41$\pm$1.47 \\ \bottomrule
\end{tabular}
\caption{RAG setting scores for FALSE passages. In the RAG framework, all scores for FALSE passages are significantly lower compared to TRUE passages, indicating that models have some ability to differentiate between correct and incorrect information. This suggests that RAG enhances factuality awareness, leading to a more discerning evaluation of misinformation.}
\label{tab:rag-false score}
\end{table*}

\section{Pointwise Reranking Phase Results}
\label{sec:pointwise-raw}
To provide a comprehensive understanding of our results in Section~\ref{sec: pointwise_reranking}, Tables~\ref{tab:normal-true score} through Table~\ref{tab:rag-false score} presents the original score distribution of pointwise reranking phase.

\section{Passage Examples}
Table~\ref{tab:passage-example} shows an example of human, GPT-3.5, GPT-4o-mini, Gemini, LLaMA and Mistral authored passages.  The figure shows an example of text comparison, highlighting that while the information contained in both model-generated passages mirrors that of the human passage, the style of model-generated text differs from that of human writing.

\begin{table*}
    \centering\small
    \begin{tabular}{p{15cm}} 
        \toprule
        \bf Generated Passage Example \\
        \midrule
        \it Human \\
        The Hemisphere. The Pantheon is a building in Rome which was originally built as a temple to the seven deities of the seven planets in the Roman state religion, but which has been a Christian church since the 7th century. \\
        \midrule
        \it GPT-3.5 \\
        The Pantheon, located in Rome, was initially constructed as a temple dedicated to the seven deities representing the seven planets in the Roman state religion. However, it has served as a Christian church since the 7th century. \\
        \midrule
        \it GPT-4o-mini \\
        The Pantheon, located in Rome, is an architectural marvel that was initially constructed as a temple dedicated to the seven deities associated with the seven planets of ancient Roman religion. However, since the 7th century, it has served as a Christian church. \\
        \midrule
        \it Gemini\\
        Picture this: the Pantheon, a Roman marvel in the heart of Rome. It began its life as a grand temple, dedicated to the celestial seven \u2013 the deities associated with the planets that ruled the Roman cosmos. But fast forward to the 7th century, and this pagan sanctuary was reborn, transformed into a Christian church, a testament to the shifting tides of faith and the enduring power of architectural reuse. \\
        \midrule
        \it LLaMA \\
        The Pantheon in Rome, initially constructed as a temple to honor the seven celestial deities of the Roman state's pantheon, has undergone a significant transformation over time, evolving from a place of pagan worship to a Christian church by the 7th century. \\
        \midrule
        \it Mistral \\
        The Pantheon, a structure in Rome, was initially constructed as a temple dedicated to the seven celestial deities of the Roman state religion. However, it transitioned into a Christian church during the 7th century, preserving its grandeur while adopting a new purpose. \\
        \bottomrule
    \end{tabular}
    \caption{Comparison of passages generated by human, GPT-3.5, GPT-4o-mini, Gemini, LLaMA and Mistral.}
    \label{tab:passage-example}
\end{table*}

\end{document}